\def\year{2022}\relax
\documentclass[letterpaper]{article} 
\usepackage{aaai22}  
\usepackage{times}  
\usepackage{helvet}  
\usepackage{courier}  
\usepackage[hyphens]{url}  
\usepackage{graphicx} 
\usepackage{natbib}  
\usepackage{caption} 
\DeclareCaptionStyle{ruled}{labelfont=normalfont,labelsep=colon,strut=off} 
\usepackage{algorithm}
\usepackage{algorithmic}

\usepackage{lineno}
\usepackage{amsmath}
\usepackage{multirow}
\usepackage{array}
\usepackage{hyperref}
\usepackage{amsfonts}
\newcolumntype{C}[1]{>{\centering\arraybackslash}m{#1}}
\newcolumntype{R}[1]{>{\raggedleft\arraybackslash}m{#1}}
\newcolumntype{P}[1]{>{\raggedright\arraybackslash}p{#1}}
\newcolumntype{M}[1]{>{\centering\arraybackslash}m{#1}}

\usepackage{newfloat}
\usepackage{listings}
\floatstyle{ruled}
\newfloat{listing}{tb}{lst}{}
\floatname{listing}{Listing}

\title{Delving into Probabilistic Uncertainty for \\ Unsupervised Domain Adaptive Person Re-identification}

\author {
    Jian Han, 
    Ya-Li Li\thanks{Corresponding Author},
    Shengjin Wang 
}
\affiliations {
    Beijing National Research Center for Information Science and Technology (BNRist) \\
    Department of Electronic Engineering, Tsinghua University \\
    hanj19@mails.tsinghua.edu.cn, \{liyali13, wgsgj\}@tsinghua.edu.cn
}

\usepackage{bibentry}

\begin{document}

\maketitle

\begin{abstract}
	Clustering-based unsupervised domain adaptive (UDA) person re-identification (ReID) reduces exhaustive annotations. However, owing to unsatisfactory feature embedding and imperfect clustering, pseudo labels for target domain data inherently contain an unknown proportion of wrong ones, which would mislead feature learning. In this paper, we propose an approach named probabilistic uncertainty guided progressive label refinery (P$^2$LR) for domain adaptive person re-identification. First, we propose to model the labeling uncertainty with the probabilistic distance along with ideal single-peak distributions. A quantitative criterion is established to measure the uncertainty of pseudo labels and facilitate the network training. Second, we explore a progressive strategy for refining pseudo labels. With the uncertainty-guided alternative optimization, we balance between the exploration of target domain data and the negative effects of noisy labeling. On top of a strong baseline, we obtain significant improvements and achieve the state-of-the-art performance on four UDA ReID benchmarks. Specifically, our method outperforms the baseline by \emph{6.5\%} mAP on the Duke2Market task, while surpassing the state-of-the-art method by \emph{2.5\%} mAP on the Market2MSMT task. Code is available at: \url{https://github.com/JeyesHan/P2LR}.
\end{abstract}

\section{Introduction}

Person re-identification (ReID) aims to retrieve all images of the target person from non-overlapping camera views. It has broad applications in smart retail, searching missing children, and other person related scenarios. Although existing deep learning methods \cite{zhang2021person, isobe2020intra, luo2019strong, guo2019beyond, hermans2017defense} have achieved remarkable performance, these methods rely heavily on manual annotations. Furthermore, they fail to generalize on other datasets when domain gap between the source and target domain exists. To address this critical issue, UDA person ReID  where labeled source images and unlabeled target images are presented has attracted great attention in recent years.

\begin{figure}[t]
	\center{\includegraphics[width=0.47\textwidth]
		{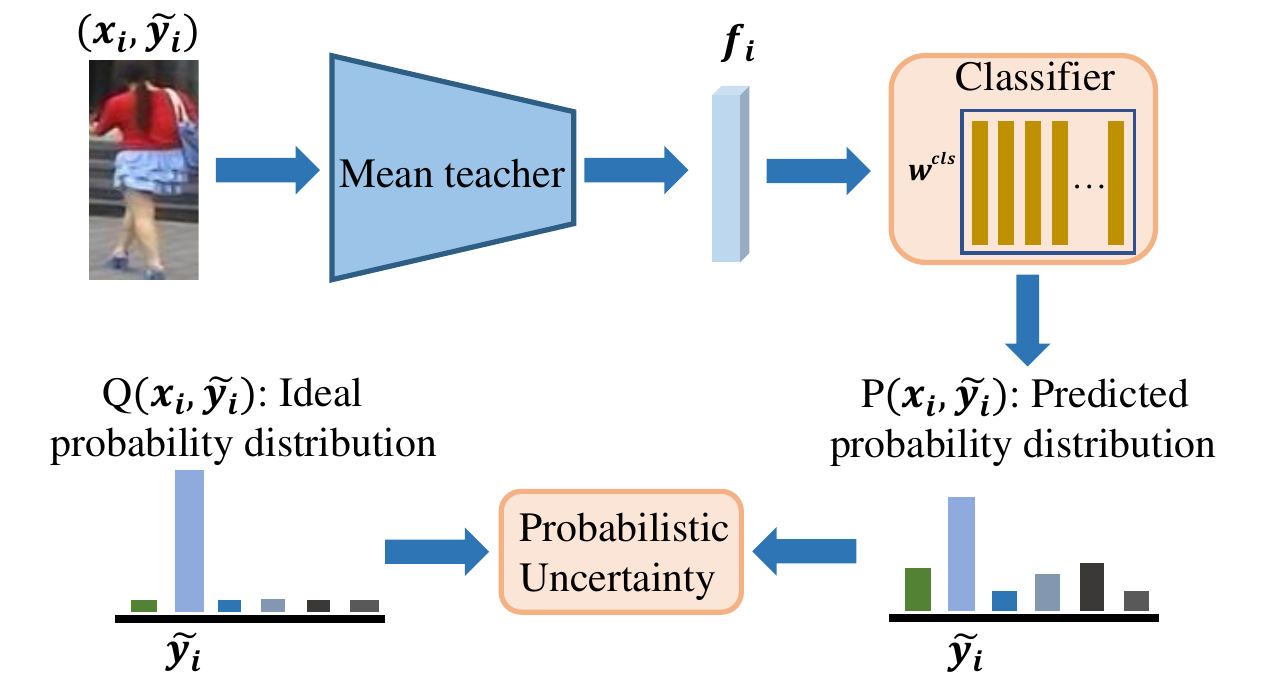}}
	\caption{\label{struct}Probabilistic uncertainty in P$^2$LR. For a target domain image $x_i$ with pseudo label $\tilde{y}_i$, we extract its feature $f_i$  and obtain the predicted distribution $P{(x_i, \tilde{y}_i)}$. We estimate the uncertainty of $(x_i, \tilde{y}_i)$ by calculating the KL probabilistic distance with the ideal distribution $Q(x_i, \tilde{y}_i)$.}
	\label{fig:correctness}
\end{figure}

UDA person ReID methods include domain transfer based methods~\cite{huang2020real, ge2020structured,deng2018image,wei2018person,zhong2018generalizing}, ranking based methods~\cite{zhong2019invariance,wang2020cross,yu2019unsupervised}, and clustering based methods~\cite{fan2018unsupervised,lin2019bottom,fu2019self,ge2020mutual,zhai2020multiple}. In general, clustering based methods can obtain superior performance. Typically, clustering based methods consist of three stages, \emph{i.e.}, i) pre-training with labeled source domain images, ii) generating pseudo labels for target domain images by clustering, and iii) fine-tuning with target domain images. The first stage is conducted only once and the latter two stages are repeated several times for mutual promotion. However, owing to unsatisfactory feature embedding and imperfect clustering quality, we are supposed to tackle an unknown proportion of wrong pseudo labels. On one hand, such noisy labels inherent with cross-domain pseudo labeling would mislead the network optimization in the fine-tuning stage. On the other hand, the negative effects caused by wrong pseudo labels will propagate and amplify as the training procedure proceeds. Therefore, it is crucial to identify wrong pseudo labels and alleviate the uncertainty of pseudo labeling to promote the UDA person ReID.

This paper addresses two key issues for clustering-based methods: 1) how to identify the wrong pseudo labels. 2) how to reduce the negative effects of wrongly-labeled samples in optimization. For the first issue, we observe that the probability distribution among identities of wrong-labeled samples is significantly different from that of samples with correct pseudo labels. Samples with correct pseudo labels are explicit, reaching high in the corresponding true identity and remain close to zero in other identities. In contrast, samples with wrong pseudo labels are ambiguous, with several peaks in several other identities. This obvious difference inspires us to model the probabilistic uncertainty by measuring the inconsistency between the predicted and ideal distributions of pseudo labels (Figure~\ref{fig:correctness}). For the second issue, we propose a probabilistic uncertainty guided progressive label refinery framework and solve it by Alternative Convex Search \cite{gorski2007biconvex}. We alternatively include samples with high credibility and minimize the uncertainty of selected samples to boost the performance of domain adaptive person ReID. 

Our main contributions can be summarized as follows: 
\begin{itemize}
	\item We propose to measure the probabilistic uncertainty of pseudo labels for UDA person ReID. A quantitative criterion that estimates the probabilistic distance of the predicted distribution of samples against ideal ones, is established to serve as the standard for eliminating wrongly-labeled samples.  
	\item We propose a framework named Probabilistic uncertainty guided Progressive Label Refinery (P$^2$LR) to determine and purify wrong labels of target domain samples for UDA person ReID.
	\item We conduct extensive experiments and achieve the state-of-the-art performance on four benchmark datasets. P$^2$LR outperforms the baseline by \emph{6.5\%} mAP on the Duke2Market task and surpasses the state-of-the-art method by \emph{2.5\%} mAP on the Market2MSMT task.
\end{itemize} 

\begin{figure*}[t]
	\center{\includegraphics[width=0.9\textwidth]
		{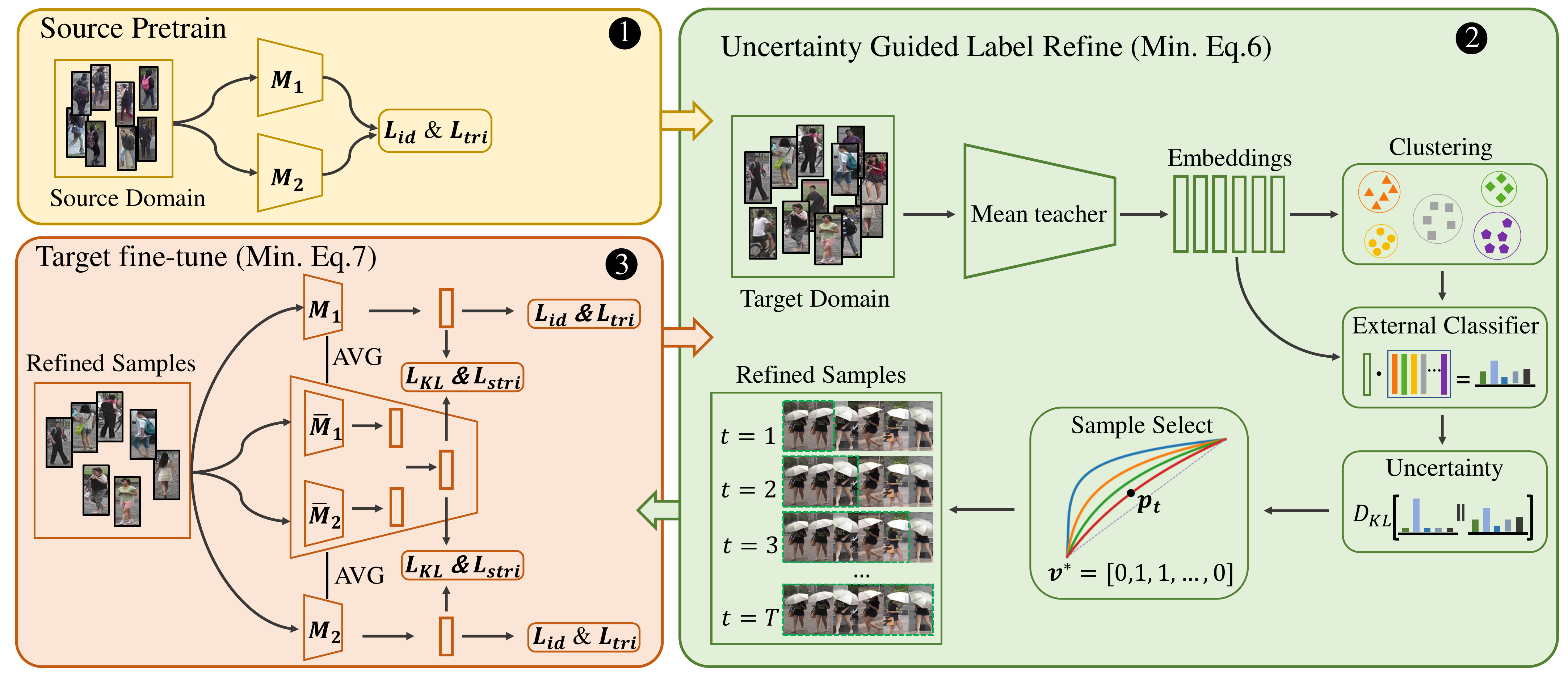}}
	\caption{\label{struct}The overall architecture of Probabilistic uncertainty guided Progressive Label Refinery (P$^2$LR) framework. It includes soure pretraining, uncertainty guided label refinery, and model fine-tuning. The latter two stages are alternatively optimized.}
	\label{fig:framework}
\end{figure*}

\section{Related Work}
\textbf{UDA person ReID. }   To save exhaustive annotations, unsupervised domain adaptation (UDA) \cite{wei2021metaalign,isobe2021multi,cui2020gradually,gong2019dlow} makes models trained on source domain adapt to target domain. Existing UDA ReID methods can be divided into three categories, \emph{i.e.}, domain transfer based methods, ranking based methods, and clustering based methods.

Domain transfer based methods \cite{huang2020real, ge2020structured} leverage style-transfer techniques to transfer source images into target domain. Then the transferred images with inherited labels are utilized to fine-tune the model pre-trained on source domain. SPGAN \cite{deng2018image} and PTGAN \cite{wei2018person} use GANs to transform source-domain images to match the image styles of the target domain. HHL \cite{zhong2018generalizing} proposes to learn camera-invariant embeddings via transforming images according to target camera styles. However, the image generation quality is still unsatisfactory and the information of target domain is not fully explored because target domain images are only used for providing style supervision signals.

Ranking based methods build positive and negative sample pairs and use the contrastive loss to optimize neural networks with soft labels on the target domain. ECN \cite{zhong2019invariance} builds a memory bank to store seen features. Exemplar, camera, and neighborhood invariance are modeled by pair learning with features in the memory bank. Wang \emph{et al.} \cite{wang2020cross} use memory bank to mine hard negative instances across batches. MAR \cite{yu2019unsupervised} exploits a set of reference persons to generate multiple soft labels. However, the constructed image pairs or feature pairs lacks global guidance over all identities. Additionally, the reference images or features might be outliers  in some cases which mislead the feature learning.

Clustering based methods obtain superior performance to domain transfer based and ranking based methods to date. Clustering methods generate pseudo labels through clustering and then fine-tune models with generated pseudo labels. Fan \emph{et al.} \cite{fan2018unsupervised} propose to alternatively assign labels for unlabeled training samples and optimize the network with the generated targets. BUC \cite{lin2019bottom} proposes a bottom-up clustering framework to gradually group similar clusters. SSG \cite{fu2019self} employs both global body and local body part features for clustering and evaluation.  DAAM \cite{Huang2020aaai} introduces domain alignment constraints and an attention module.  MMT \cite{ge2020mutual} introduces two sibling mutual mean teachers. MEB-Net \cite{zhai2020multiple} establishes three networks to perform mutual mean learning. However, these methods ignore noisy/wrong labels generated by clustering methods, which hinders the advancement of these approaches. 

\noindent\textbf{Pseudo label evaluation. }   Wrong pseudo label samples harm learning robust feature embedding for neural networks. To quantify and identify the correctness of pseudo labels, pseudo label evaluation becomes significantly crucial.  Kendall \emph{et al.} \cite{kendall2017uncertainties} and Chang \emph{et al.} \cite{chang2020data} establish an end-to-end framework to measure the observation noise and alleviate the negative influence for better network optimization. Zheng \emph{et al.} \cite{zheng2021rectifying} propose to estimate the correctness of predicted pseudo labels in semantic segmentation. As for clustering-based UDA person ReID, EUG \cite{wu2019progressive}, UNRN \cite{zheng2021exploiting}, and GLT \cite{zheng2021group} are uncertainty based methods. EUG  employs the $l_2$ distance between samples and clustering centroids in feature space to determine the reliability of samples. UNRN  measures the output consistency between mean teacher and student models as uncertainty, while we model the distribution of single-model clusters and measure the probabilistic distance to an ideal distribution as the uncertainty. GLT  is a group-aware label transfer framework to explicitly correct noisy labels while we pick reliable pseudo labels to train the model progressively, which further correct noisy labels implicitly.

\section{Methodology}
\textbf{Notation.} The goal of UDA person ReID is to adapt the trained model from a source domain $\mathbb{D}_s=\{(x^s_i, y^s_i)|^{N_s}_{i=1}\}$ to an unsupervised target domain $\mathbb{D}=\{x_i|^{N}_{i=1}\}$, where $N_s$ and $N$ denote the number of samples in the source and target domains, respectively. $x^s_i$ and $y^s_i$ denote the sample and its attached label in the source domain with supervisory information. $x_i$ denotes the sample in target domain without supervision. The pseudo label generated by clustering for sample $x_i$ in target is denoted as $\tilde{y}_i$.

Figure~\ref{fig:framework} shows an overview of our proposed P$^2$LR framework for UDA person ReID. We first construct a clustering baseline with mutual mean teachers \cite{ge2020mutual}. On top of the baseline, we introduce the probabilistic uncertainty guided progressive label refinery to evaluate the noisy level of pseudo labels and reduce the negative influence of noisy samples. In the following parts, we first introduce the clustering baseline, and then elaborate on the label uncertainty modeling based on probabilistic uncertainty and the proposed P$^2$LR framework.

\subsection{Recap Clustering Baseline} 
We construct a clustering-based pipeline based on MMT \cite{ge2020mutual} for UDA person ReID. Following the general pipeline of clustering-based methods, three stages are consisted, such as: source-domain model pre-training, clustering, and target-domain fine-tuning. In the source pre-training stage, two sibling networks are established with the same architecture. They are initialized with different random seeds, trained with the same data yet experienced different data augmentations to reduce the dependence. The networks are optimized by the identity loss $L_{id}$ and triplet loss $L_{tri}$. In the clustering stage, pseudo labels for target-domain data are generated by clustering. In the target fine-tuning stage, two mean teachers $\bar{M_1}$ and $\bar{M_2}$ are established with the exponential moving average of student models over iterations. The prediction of one teacher model is taken as soft labels to mutually supervise the training of the other student. Apart from the identification and triplet loss with hard pseudo labels provided by clustering, each model is also guided by the Kullback–Leibler (KL) divergence loss $L_{KL}$ and soft triplet loss $L_{stri}$ with the soft labels.

The whole loss function in the fine-tuning stage is as:
\begin{equation}
	\begin{aligned}
		\label{eq:loss}
		\mathcal{L} = \mathcal{L}_{id} + \lambda_{tri} \mathcal{L}_{tri} + \lambda_{KL} \mathcal{L}_{KL} + \lambda_{stri} \mathcal{L}_{stri}
	\end{aligned}
\end{equation}
where $\lambda_{tri}$, $\lambda_{KL}$, $\lambda_{stri}$ indicate corresponding loss weights.

\begin{figure*}[t]
	\center{\includegraphics[width=0.90\textwidth]
		{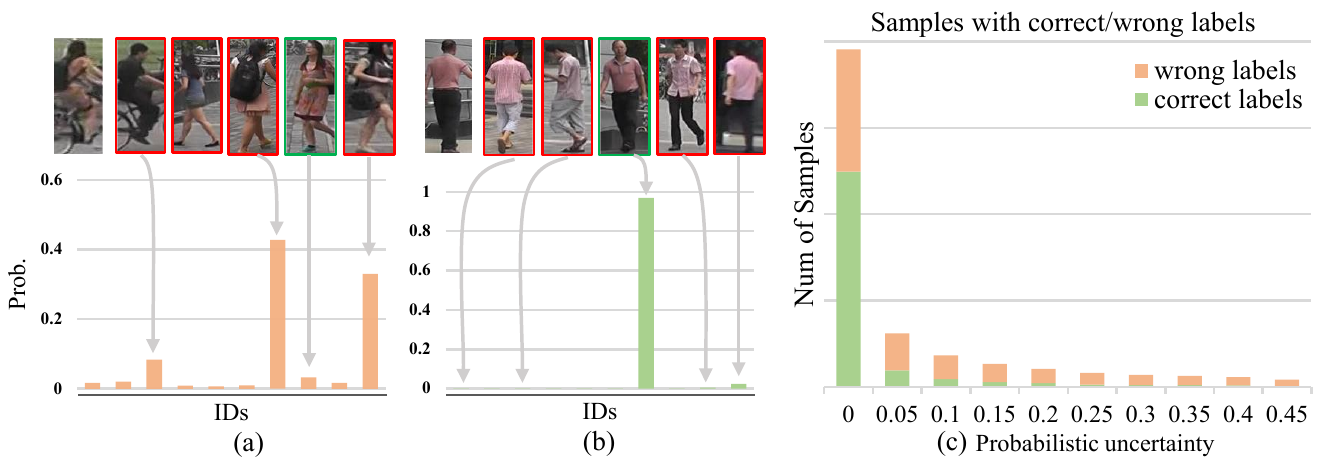}}
	\caption{\label{struct}Probability distributions of the sample with a wrong pseudo label (a) and with a correct pseudo label (b) in the first refinery step for Duke2Market. The sample with a wrong label is ambiguous, with several peaks. In contrast, the sample with a correct pseudo label is explicit, reaching high in the true identity. (c) The histogram of samples with correct/wrong labels. Our probabilistic uncertainty is highly related to correctness. Samples with low uncertainty are more likely to have correct labels.}
	\label{fig:statistic}
\end{figure*}

\subsection{Probabilistic Uncertainty Modeling}
For clustering-based UDA person ReID, the pseudo labels are noisy, which would mislead the network training in the target fine-tuning stage and hurt the performance in target domain. To reduce the effects of noise, uncertainty estimation is a natural way to eliminate the unreliable pseudo labels.
To establish the uncertainty criterion, we present the output probabilistic distribution of samples with correct and wrong labels for an investigative study. As shown in Figure \ref{fig:statistic}a, the sample with wrong pseudo labels is associated with high probabilistic uncertainty, which has several peaks and generally reaches high in different identities. In contrast, the sample with correct pseudo labels is associated with low probabilistic uncertainty (Figure \ref{fig:statistic}b). The assigned probability peaks high in the correct identity and remains nearly zero in other identities. This is consistent with the intuitive idea that the samples with wrong pseudo labels are ambiguous for model prediction, which is significantly distinct to correct pseudo labels.

Inspired by this observation, we leverage the distribution difference as the probabilistic uncertainty to softly evaluate the noisy level of samples. Each unlabeled sample $x_i$ in the target domain is assigned with a pseudo label $\tilde{y}_i$ by clustering. Based on the clustering at $t$-th step, we build an external classifier $\phi_t(\cdot|w^{cls}_t)$, where $w^{cls}_t \in \mathbb{R}^{c \times d}$ is the parameterized weights of the external classifier. Note that the weights are generated dynamically and don't need separate training. Here we employ $c$ cluster centroids of $d$ dimensions as classifier weights. For a feature $f_i \in \mathbb{R}^d$ of sample $(x_i, \tilde{y}_i)$ extracted by the mean teacher, we obtain classified probability distributions among identities with this classifier following Eq.\ref{eq:classifier}. Here $\alpha$ is the temperature parameter. 

\begin{equation}
	\label{eq:classifier}
	P{(x_i, \tilde{y}_i)} = \phi _t (f_i) = Softmax\left(\frac{w^{cls}_t}{||w^{cls}_t||} \cdot \frac{f_i}{||f_i||} \cdot \alpha\right)
\end{equation}

In source domain, we obtain a single impulse distribution for samples after symmetrically arranging the probability from large to small while making $y^s_i$ as the center. This phenomenon motivates us to model the generalized ideal distributions for samples in the target domain. We find that the smoothed $\delta$ distribution associative with large temperature $\alpha$ is highly stable and insensitive to specific datasets. When $\alpha$ decreases ($\alpha$ controls the variety of distributions), another distribution might be better for some cases but hard to generalize. Therefore, we draw the smoothed $\delta$ distribution (Eq.\ref{eq:qfunc}) as the ideal distribution $Q(x_i, \tilde{y}_i)$. 

\begin{equation}
	Q(x_i, \tilde{y}_i) = \delta_{smooth} (j-\tilde{y}_i) =\left\{
	\begin{aligned}
		\label{eq:qfunc}
		& \quad \epsilon , \quad if ~~j=\tilde{y}_i\\
		& \frac{1-\epsilon}{c-1} , otherwise\\
	\end{aligned}
	\right.
\end{equation}
where $j$ is the identity index and $c$ is the number of identities (\emph{i.e.} the number of clusters). $\epsilon$ is a hyperparameter and is set to 0.99.

We estimate the probabilistic distance of the predicted probability across identities and the ideal distribution to measure the uncertainty of samples. Rather than measuring the feature inconsistency of teacher-student models~\cite{zheng2021exploiting}, we define a criterion named \textit{probabilistic uncertainty}, which measures the inconsistency between predicted distribution $P{(x_i, \tilde{y}_i)}$ and the ideal distribution $Q(x_i, \tilde{y}_i)$. Moreover, we leverage the Kullback–Leibler (KL) divergence to measure the inconsistency and establish the probabilistic uncertainty as:

\begin{equation}
	\label{eq:uncertainty}
	U ~ \left(x_i, \tilde{y}_i\right)= D_{KL}\left(Q\left(x_i, \tilde{y}_i\right)||P(x_i, \tilde{y}_i)\right)
\end{equation}

Larger $U ~(x_i, \tilde{y}_i)$ indicates that the pseudo label generated by clustering is more likely to be wrong, which should be excluded in the target fine-tuning stages. As in Figure \ref{fig:statistic}c, our probabilistic uncertainty is highly related to correctness.

\subsection{Uncertainty Guided Alternative Optimization} 

We formulate the uncertainty guided domain adaption mathematically. The model can be denoted as $\emph{M}(\cdot|\mathbf{w})$, where $\mathbf{w}$ is the associated parameters. Each model maps a sample $x_i$ to the prediction $M\left(x_i|\mathbf{w}\right)$. An uncertainty score can be further obtained, as $u_i=U(x_i, \tilde{y}_i|\mathbf{w})$. $U$ is the label uncertainty estimation module. The uncertainty guided learning can be formulated as a joint optimization problem as: 
\begin{equation}
	\begin{aligned}
		\label{eq:target}
		\min_{\mathbf{w};\mathbf{v}\in{\{0,1\}}^N} \mathbb{E}(\mathbf{w},\mathbf{v};\beta) =\sum_{i=1}^{N} v_i u_i -\beta \sum_{i=1}^{N} v_i
	\end{aligned}
\end{equation}
where $v_i \in{\{0,1\}}$ is the indicator for sample selection. $\beta$ is the age parameter to control the learning pace. That is, we need to reduce the overall uncertainty of select samples  while selecting as many samples as possible for sufficient training. To optimize the above objective function, we adopt Alternative Convex Search \cite{gorski2007biconvex}. In particular, $\mathbf{w}$ and $\mathbf{v}$ are alternatively optimized while fix the other. With the fixed  $\mathbf{w}^*$, the global optimum $\mathbf{v}^*$ is solved as:

\begin{equation}
	\begin{aligned}
		\label{eq:vstep}
		\mathbf{v}^* = \arg \min_{v_i \in \{0, 1\}} \sum_{i=1}^{N} v_i u_i - \beta \sum_{i=1}^{N} v_i
	\end{aligned}
\end{equation}

When $\mathbf{v}^*$ is fixed, the global optimum $\mathbf{w}^*$ is solved as:

\begin{equation}
	\begin{aligned}
		\label{eq:wstep}
		\mathbf{w}^* = \arg \min_{\mathbf{w}} \sum^N_{i=1} v^*_i u_i
	\end{aligned}
\end{equation}

The two optimization steps are iteratively conducted, while $\beta$ is gradually increased to add more harder samples.

\textbf{Uncertainty guided sample selection.} 
With the proposed uncertainty measurement, the solution for Eq.\ref{eq:wstep} is provided by gradient descent based network training. With fixed $\mathbf{w}^*$, finding the optimal $\mathbf{v}$ is converted into a combinatorial optimization issue with the linear objective function. We can simply derive the global minimum by setting the partial derivative of Eq.\ref{eq:vstep} to $v_i$ as zero. Considering $v_i$ is either 0 or 1, we obtain the close-formed solution $\mathbf{v}^*$ as:

\begin{equation}
	v^*_i =\left\{
	\begin{aligned}
		\label{eq:vderive}
		1 ,~~if ~~u_i \leq \beta \\
		0 ,~~otherwise\\
	\end{aligned}
	\right.
\end{equation}

Since clustering and target fine-tuning stages are iterated in an alternative way for multiple steps, we need to determine how many samples are included. A sequence $\{N \cdot p_t\}, t=0,1,\cdots, T$ is predefined for this purpose, where $N \cdot p_t$ is the number of selected examples at iteration time $t$. Based on the predefined $p_t$, the threshold $\beta$ is dynamically updated according to Eq.\ref{eq:lambda_t} to ensure exactly $N \cdot p_t$ samples are assigned with non-zero $v_i$. 

\begin{equation}
	\label{eq:lambda_t}
	\beta = Sort \{u_1, u_2, ..., u_N\}_{(N\cdot p_t)}
\end{equation}

Note that as the training proceeds, the models become more reliable. Consequently, the measured uncertainty reduces drastically at the beginning, then converges slowly with larger $t$ (Figure~\ref{fig:pt}(left)). As the result, $p_t$ should be enlarged as $t$ increases. To be consistent with the trends of uncertainty reduction, we design $p_t$ as a rescaled logarithm-exponential function:

\begin{figure}[b]
	\center{\includegraphics[width=0.47\textwidth]
		{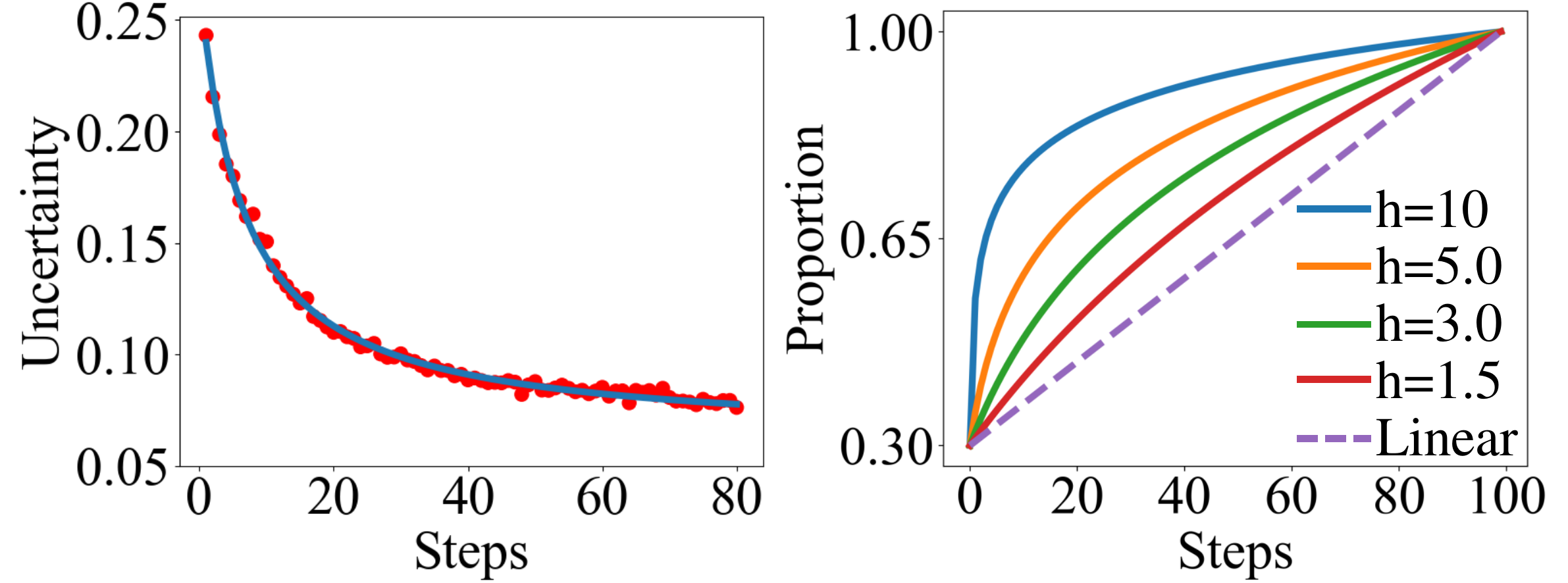}}
	\caption{Left: The curve of mean uncertainty against training steps on Market dataset. Right: $p(t)$ with different $h$.}
	\label{fig:pt}
\end{figure}

\begin{equation}
	\label{eq:ln}
	p_t = p(t) = p_0 + \frac{1}{h} \ln \left(1 + \left(e^{h \cdot (1-p_0)}-1\right) \cdot \frac{t}{T}\right)
\end{equation}
$p_0$ is the initial proportion of samples for fine-tuning in target domain. The setting of $p_0$ enables the model to learn from sufficient easy samples at early steps, to avoid falling into the negative loop of degrading. $h$ is the hyper-parameter to control the increasing rate of samples (Figure~\ref{fig:pt}(right)). Because the rate of adding examples is negatively correlated to the size of the current training set, the number of newly added examples is reduced as the training proceeds. More importantly, the uncertainty decreases dramatically when $t$ is small, we enlarge the set of samples in a fast way. As $t$ increases, the uncertainty converges and the set of samples for fine-tuning increases slowly for stable training. This progressive label refinery would help the model learn sufficiently from the newly added samples in target domain.

\begin{table*}[t!]
	\centering
	\begin{center}
		\begin{tabular}{|P{6.8cm}|C{0.8cm}C{0.8cm}C{0.8cm}C{0.8cm}|C{0.8cm}C{0.8cm}C{0.8cm}C{0.8cm}|}
			\hline
			\multicolumn{1}{|c|}{\multirow{2}{*}{Methods}} & \multicolumn{4}{c|}{DukeMTMC$\to$Market1501} & \multicolumn{4}{c|}{Market1501$\to$DukeMTMC} \\
			\cline{2-9}
			\multicolumn{1}{|c|}{} & mAP & R1 & R5 & R10 & mAP & R1 & R5 & R10 \\
			\hline
			ATNet~\cite{liu2019adaptive}(CVPR'19) & 25.6& 55.7& 73.2& 79.4& 24.9 &45.1 &59.5& 64.2\\
			SPGAN+LMP~\cite{deng2018image}(CVPR'18) &26.7 &57.7 &75.8 &82.4& 26.2&46.4 &62.3 &68.0 \\
			BUC~\cite{lin2019bottom} (AAAI'19) & 38.3 & 66.2 & 79.6 & 84.5 & 27.5 & 47.4 & 62.6 & 68.4 \\
			ECN~\cite{zhong2019invariance} (CVPR'19) & 43.0 & 75.1 & 87.6 & 91.6 & 40.4 & 63.3 & 75.8 & 80.4 \\
			PDA-Net~\cite{li2019cross} (ICCV'19) & 47.6 & 75.2 & 86.3 & 90.2 & 45.1 & 63.2 & 77.0 & 82.5 \\
			PCB-PAST~\cite{zhang2019self} (ICCV'19) & 54.6 & 78.4 & - & - & 54.3 & 72.4 & - & - \\
			SSG~\cite{fu2019self} (ICCV'19) & 58.3 & 80.0 & 90.0 & 92.4 & 53.4 & 73.0 & 80.6 & 83.2 \\
			ACT~\cite{yang2020asymmetric} (AAAI'20) & 60.6 & 80.5 & - & - & 54.5 & 72.4 & - & - \\
			MPLP~\cite{wang2020unsupervised} (CVPR'20) & 60.4 &84.4 &92.8& 95.0 & 51.4&72.4 &82.9& 85.0  \\
			DAAM~\cite{Huang2020aaai} (AAAI'20)& {67.8} & {86.4} & {-} & {-} & {63.9} & {77.6} & {-} & {-} \\ 
			AD-Cluster~\cite{zhai2020ad} (CVPR'20)& {68.3} & {86.7} & {94.4} & {96.5} & {54.1} & {72.6} & {82.5} & {85.5} \\
			MMT~\cite{ge2020mutual} (ICLR'20) & {71.2} & {87.7} & {94.9} & {96.9} & {65.1} & {78.0} & 88.8 & 92.5 \\
			NRMT~\cite{zhao2020unsupervised}(ECCV'20) & 71.7& 87.8& 94.6& 96.5& 62.2& 77.8& 86.9& 89.5 \\
			B-SNR+GDS-H~\cite{jin2020global}(ECCV'20) & 72.5 & 89.3 & - & - & 59.7 & 76.7 &- &-\\
			MEB-Net~\cite{zhai2020multiple}(ECCV'20) & 76.0 &89.9& 96.0& 97.5& 66.1& 79.6 & 88.3& 92.2 \\
			UNRN~\cite{zheng2021exploiting} (AAAI'21) & 78.1 & 91.9 & 96.1 & 97.8 & 69.1 & 82.0 & 90.7 & 93.5 \\
            GLT~\cite{zheng2021group}(CVPR'21) & 79.5 &92.2& 96.5& 97.8& 69.2& 82.0 & 90.2& 92.8 \\
			\hline
			P$^2$LR (Ours) & \textbf{81.0} & \textbf{92.6} & \textbf{97.4} & \textbf{98.3} & \textbf{70.8} & \textbf{82.6} & \textbf{90.8} & \textbf{93.7} \\

			\hline
			
		\end{tabular}\\

		\begin{tabular}{|P{6.8cm}|C{0.8cm}C{0.8cm}C{0.8cm}C{0.8cm}|C{0.8cm}C{0.8cm}C{0.8cm}C{0.8cm}|}
			\hline
			\multicolumn{1}{|c|}{\multirow{2}{*}{Methods}} & \multicolumn{4}{c|}{Marke1501$\to$MSMT17} & \multicolumn{4}{c|}{DukeMTMC$\to$MSMT17} \\
			\cline{2-9}
			\multicolumn{1}{|c|}{} & mAP & R1 & R5 & R10 & mAP & R1 & R5 & R10 \\
			\hline
			ECN~\cite{zhong2019invariance} (CVPR'19) & 8.5 & 25.3 & 36.3 & 42.1 & 10.2 & 30.2 & 41.5 & 46.8 \\
			SSG~\cite{fu2019self} (ICCV'19) & 13.2 & 31.6 &- & 49.6 & 13.3 & 32.2 & - & 51.2 \\
			DAAM~\cite{Huang2020aaai} (AAAI'20)& {20.8} & { 44.5} & {-} & {-} & { 21.6} & { 46.7} & {-} & {-} \\
			NRMT~\cite{zhao2020unsupervised}(ECCV'20) & 19.8& 43.7& 56.5 &62.2& 20.6 &45.2& 57.8& 63.3 \\
			MMT~\cite{ge2020mutual} (ICLR'20) & 22.9 & 49.2 & 63.1 & 68.8 & 23.3 & 50.1 & 63.9 & 69.8 \\
			UNRN~\cite{zheng2021exploiting} (AAAI'21)  & 25.3 & 52.4 & 64.7 & 69.7 & 26.2 & 54.9 & 67.3 & 70.6 \\
            GLT~\cite{zheng2021group}(CVPR'21) & 26.5 &56.6& 67.5& 72.0& 27.7& 59.5 & 70.1& 74.2 \\
			\hline
			P$^2$LR (Ours) & \textbf{29.0} & \textbf{58.8} & \textbf{71.2} & \textbf{76.0} & \textbf{29.9} & \textbf{60.9} & \textbf{73.1} & \textbf{77.9} \\
			\hline
			
			\hline
		\end{tabular}
	\end{center}
	\caption{Performance (\%) comparison with the state-of-the-art methods for UDA person ReID on the datasets of DukeMTMC-reID, Market-1501 and MSMT17. Results of the best are marked by bold text.
	}
	\label{tab:sota}
\end{table*}

\begin{table}[h]
	\centering
	\begin{center}
		\begin{tabular}{P{4.0cm}|C{0.55cm}C{0.55cm}C{0.55cm}C{0.5cm}}
			\hline
			\multicolumn{1}{c|}{\multirow{2}{*}{Methods}} & \multicolumn{4}{c}{Duke$\to$Market}   \\
			\cline{2-5}
			\multicolumn{1}{c|}{}& mAP & R1 & R5 & R10 \\
			\hline
			Model pretraining      &29.6&57.8&73.0&79.0\\
			\hline
			Base.               &58.0&78.1&89.0&92.2\\
			Base.+ML      &68.5&84.6&94.3&96.1\\
			Base.+ML+MT (Baseline) &74.5&90.3&96.4&97.9\\
			\hline
			Baseline + $l_2$ distance &77.8&90.8&96.6&98.0\\
			Baseline + Internal classifier &80.2&91.9&96.9&98.0\\
			Baseline + Consistency &79.2&90.4&96.8&98.0\\
			Baseline + Reweighting          &77.7&90.6&96.6&97.6 \\
			Baseline + P$^2$LR($\epsilon$=0.95) &74.8&89.9&95.7&97.1\\
			Baseline + P$^2$LR($\epsilon$=0.97) &77.1&90.0&96.4&97.5\\
			Baseline + P$^2$LR($\epsilon$=1.00) &80.8&91.8&97.4&98.2\\
			Baseline + P$^2$LR (Ours)           &\textbf{81.0}&\textbf{92.6}&\textbf{97.4}&\textbf{98.3}\\
			\hline
		\end{tabular}
	\end{center}
	\caption{Ablation studies on the effectiveness of components in the proposed method on the Duke2Market benchmark.}
	\label{tab:ablation1}
\end{table}

\section{Experiments}
\subsection{Implementation Details}
We implement our model based on MMT~\cite{ge2020mutual} and train it on four Tesla XP GPUs. ADAM optimizer is adopted to optimize models with the weight decay of 5e-4. We employ ImageNet \cite{deng2009imagenet} pre-trained ResNet50 \cite{he2016deep} as the backbone. We obey normal $P-K$ sampling in person ReID, where $P=4$ and $K=16$ in each mini-batch. Then we perform data augmentation of random cropping, flipping, and erasing. Note that random erasing is not utilized in the source pre-training stage. All person images are resized to $256 \times 128$. We use the clustering algorithm of k-means where the number of clusters ($c$) is set as 500, 700, and 1500 for Market, Duke, and MSMT datasets, respectively. The temperature parameter $\alpha$ in Eq.\ref{eq:classifier} is set to 20. The parameter $p_0$ and $h$ in Eq.\ref{eq:ln} are set to 0.3 and 1.5, respectively. Overall alternative optimization steps $T$ are set to 100. In source pre-training stage, the initial learning rate is set to $3.5\times 10^{-4}$ and is decreased by 1/10 on the 40th and 70th epoch in the total 80 epochs. In target fine-tuning stage, the learning rate is fixed to $3.5\times 10^{-4}$.

\subsection{Datasets and Protocols}

We evaluate our method on three main-stream person ReID datasets, \emph{i.e.}, Market-1501 (Market) \cite{zheng2015scalable}, DukeMTMC-reID (Duke) \cite{ristani2016performance}, and MSMT17 (MSMT) \cite{wei2018person}. The Market-1501 dataset contains 32,668 annotated images of 1,501 identities shot from 6 cameras, where 12,936 images of 751 identities are used for training and 19,732 images of 750 identities for testing.The DukeMTMC-reID dataset consists of 36,411 images collected from 8 cameras, where 702 identities are used for training and 702 identities for testing. MSMT17 is the most challenging and largest person ReID dataset consisting of 126,441 images of 4,101 identities, where 1,041 identities and 3,060 identities are used for training and testing, respectively. We adopt mean average precision (mAP) and CMC Rank-1/5/10 (R1/R5/R10) accuracy without re-ranking \cite{zhong2017re} for evaluation.

\subsection{Comparison with State-of-the-Arts}
We compare our method against the state-of-the-art (SOTA) methods on four UDA ReID settings and present the results in Table \ref{tab:sota}. Among existing methods for UDA person ReID, DAAM \cite{Huang2020aaai} introduces domain alignment constraints and an attention module. SSG \cite{fu2019self}, MMT \cite{ge2020mutual}, MEB-Net \cite{zhai2020multiple}, and UNRN \cite{zheng2021exploiting} are all clustering-based methods. SSG \cite{fu2019self} employs both global body and local body part features for clustering and evaluation. We construct the baseline based on MMT \cite{ge2020mutual} which introduces mutual mean teacher for UDA person ReID. Compared to the baseline MMT~\cite{ge2020mutual}, our proposed P$^2$LR significantly improves the UDA ReID accuracy, with 9.8\%, 5.7\%, 6.1\%, and 6.6\% mAP improvements on four UDA ReID settings. Compared to MEB-Net \cite{zhai2020multiple} which establishes three networks to perform mutual mean learning, we increase the mAP by 5.0\%, 4.7\% with a simpler architecture design. Notably, UNRN  and GLT leverage source data during target fine-tuning stage and build an external support memory to mine hard pairs. Our P$^2$LR still achieves 3.7\% and 3.7\% mAP gains to UNRN, 2.5\% and 2.2\% mAP gains to GLT on the MSMT dataset. In general, our method P$^2$LR achieves the state-of-the-art performance on all datasets, which verifies the effectiveness of P$^2$LR.

\begin{figure*}[t]
	\center{\includegraphics[width=0.9\textwidth]
		{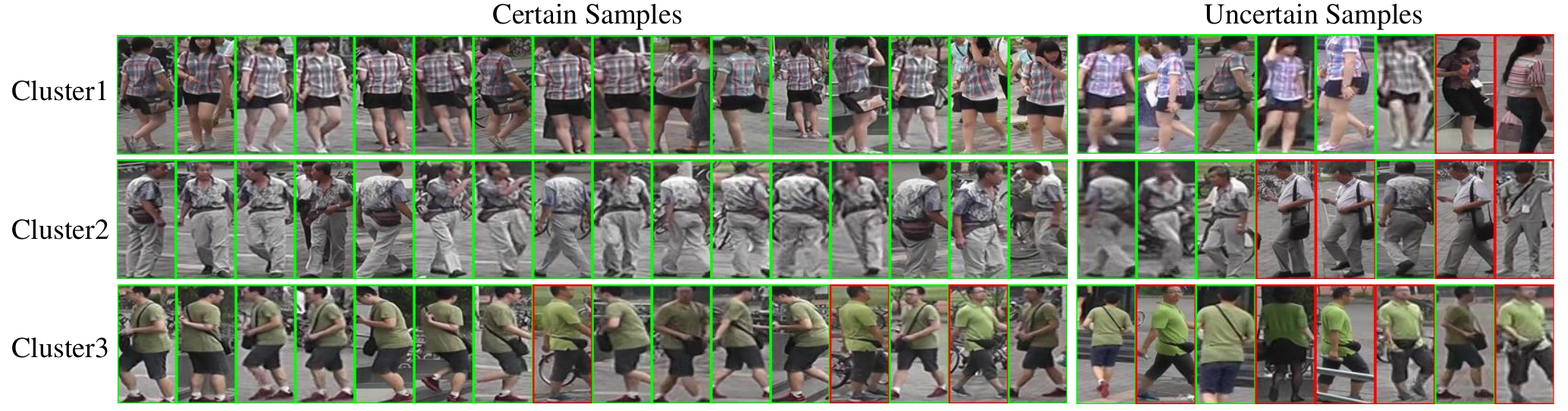}}
	\caption{\label{struct}Visualization of refined pseudo label samples. We present samples with high certainty and uncertain samples of three clusters measured in the first refinery step. We can observe that most certain samples are from the same identity while uncertain samples are hard samples of the identity or from other identities.}
	\label{fig:vis}
\end{figure*}

\begin{figure}[t]
	\center{\includegraphics[width=0.475\textwidth]
		{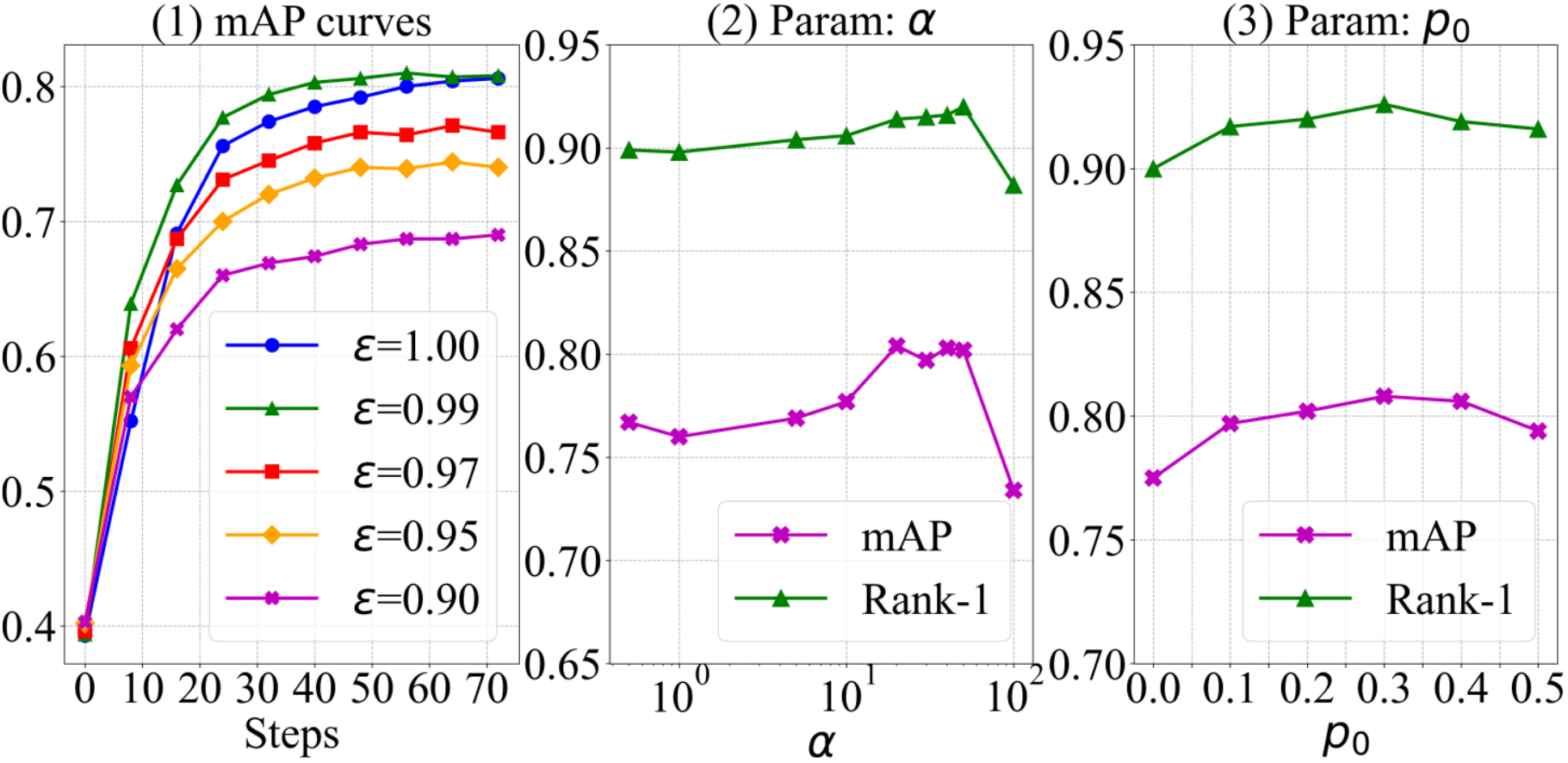}}
	\caption{Parameter analysis on Duke2Market.}
	\label{fig:param}
\end{figure}

\subsection{Ablation Studies}
In this section, we carry out extensive ablation studies to validate the effectiveness of each component in P$^2$LR.

\textbf{Effectiveness of progressive label refinery.} To alleviate the negative effects of noisy/wrong pseudo labels, we exploit the characteristics of probability distribution among identities to evaluate probabilistic uncertainty of pseudo labels and refine the pseudo labels by alternative optimization. We validate the effectiveness of the proposed progressive label refinery module by comparing the performance of \textit{Baseline} and \textit{Baseline+P$^2$LR}. Without P$^2$LR, the achieved mAP and Rank-1 accuracy are 74.5\% and 90.3\%, respectively. With P$^2$LR, the achieved mAP is 81.0\% and Rank-1 accuracy is 92.6\%, respectively. The mAP is significantly improved by 6.5\% with the sole P$^2$LR on the Duke2Market task. The results show that the proposed method is effective to alleviate the negative effects caused by noisy/wrong pseudo labels and significantly improves UDA person ReID performance. 

\textbf{Different designs of label correctness estimation. } A straightforward way to evaluate the uncertainty of pseudo labels is referring to the $l_2$ distance to the nearest centroid in feature space. As shown in Table \ref{tab:ablation1}, the mAP and Rank-1 accuracy is improved by 3.2\% and 1.8\% with the probabilistic uncertainty criterion. That is because the probability distribution based uncertainty evaluation considers more global references than $l_2$ distance in local feature space. Besides, we find introducing an external classifier with cluster centroid for uncertainty evaluation outperforms internal classifiers in $\emph{M}(\cdot|\mathbf{w})$. It can be attributed to the fact that internal classifiers are more easily misled by noisy labeling during training. External classifiers whose weights are generated by clustering could correct the misleading to some extent. 

We compare our label evaluation method to prediction consistency based method UNRN. Our method surpasses UNRN with large margins. The uncertainty (noise level) of pseudo labels should be measured from the intrinsic property of its own probabilistic distribution, especially for UDA ReID. Since noise already exists in teacher models, the consistency-based uncertainty as in UNRN might overfit biased teacher models.

\textbf{Progressive label refinery vs Sample Reweighting.} We compare the label refinery with reweighting method proposed in UNRN \cite{zheng2021exploiting}. In general, our method filters out wrong labels with the guidance of uncertainty, while the reweighting-based method reduces the negative influence in loss functions. The comparative results are presented in Table \ref{tab:ablation1}. The proposed progressive label refinery significantly outperforms reweighting method. It indicates that removing wrong labeled samples in training is more effective for reducing their negative influence.

\textbf{Parameter analysis.} We experiment with different ideal distribution $Q$ by setting different $\epsilon$. We present the mAP/R1/R5/R10 in Table \ref{tab:ablation1} and mAP curves with the number of refinery steps in Figure \ref{fig:param}a, respectively. It can be seen $\epsilon$ has important impacts while the mAP curve of $\epsilon$=0.99 grows fastest and highest. Therefore, we set $\epsilon$=0.99 for all other experiments. We then explore the influence of temperature parameter $\alpha$ in the uncertainty measurement while setting $p_0=0.2$ in Figure \ref{fig:param}b. The achieved mAP and Rank-1 accuracy maintain high when $\alpha$ is set to $20 \sim 50$. When $\alpha=20$, the achieved mAP is 80.4\% and Rank-1 accuracy is 91.4\%. When $\alpha=50$, the achieved mAP is 80.2\% and Rank-1 accuracy is 92.0\%. Afterwards, we present the influence of $p_0$ in progressive label refinery Figure \ref{fig:param}c setting $\alpha=20$. The mAP and Rank-1 accuracy reach peak (81.0\% and 92.6\%) when $p$ is around 0.3. It indicates we preserve about 30\% of samples and eliminate 70\% ones in the early refinery steps. We progressively exploit more uncertain samples in later refinery steps for strong performance.

\textbf{Visualization.} We present the visualization results to validate the effectiveness of probabilistic uncertainty measurement for selecting samples. The selected certain samples and un-selected uncertain samples of three clusters determined by the proposed criterion in the first refinery step are presented in Figure \ref{fig:vis}. It can be observed that most selected samples are from the same identity while un-selected samples are hard samples of the identity or from other identities. By combining the selected certain samples with high confidence into target domain fine-tuning, the proposed P$^2$LR can achieve superior UDA person ReID performance.

\section{Conclusion}

In this paper, we aim to find wrong pseudo labels and reduce their negative effects for clustering-based UDA person ReID. We observe that the probability distributions of samples with wrong pseudo labels are ambiguous with multiple peaks, which is different from that of correctly-labeled samples. This inspires us to propose a probabilistic uncertainty guided progressive label refinery (P$^2$LR) framework for UDA person ReID. We solve it by alternative optimization, which balances between the target domain exploitation and noise label overfitting. The proposed method brings significant improvements over a strong baseline and achieves the state-of-the-art performance on four benchmarks.

\section*{Acknowledgments}
This work was supported by the Cross-Media Intelligent Technology
Project of Beijing National Research Center for Information Science and Technology
(BNRist) under Grant No.BNR2019TD01022, National Natural Science
Foundation of China under Grant No.61771288, and the research fund under Grant No.
2019GQG0001 from the Institute of GuoQiang, Tsinghua University.

\bibliography{aaai22}

\end{document}